\documentclass[sigconf,10pt]{acmart}

\makeatletter
\def\@ACM@checkaffil{% Only warnings
    \if@ACM@instpresent\else
    \ClassWarningNoLine{\@classname}{No institution present for an affiliation}%
    \fi
    \if@ACM@citypresent\else
    \ClassWarningNoLine{\@classname}{No city present for an affiliation}%
    \fi
    \if@ACM@countrypresent\else
        \ClassWarningNoLine{\@classname}{No country present for an affiliation}%
    \fi
}
\makeatother

\pdfoutput=1
\acmSubmissionID{<PAPER ID>}
\renewcommand\footnotetextcopyrightpermission[1]{}
\AtBeginDocument{%
  }

\setcopyright{acmlicensed}
\copyrightyear{2018}
\acmYear{2018}
\acmDOI{XXXXXXX.XXXXXXX}

\acmISBN{978-1-4503-XXXX-X/18/06}
\usepackage{float}
\usepackage{url}
\usepackage{stfloats}
\usepackage{subfigure}
\usepackage{graphicx}
\usepackage{subcaption}
\usepackage{multirow}
\usepackage{algorithm}
\usepackage{algorithmic}
\begin{document}

\title{Autoware.Flex: Human-Instructed Dynamically Reconfigurable Autonomous Driving Systems}


\author{Ziwei Song}
\affiliation{%
  \institution{City University of Hong Kong}
}
\email{ziweisong5-c@my.cityu.edu.hk}

\author{Mingsong Lv}
\affiliation{%
  \institution{The Hong Kong Polytechnic University}
}
\email{mingsong.lyu@polyu.edu.hk}

\author{Tianchi Ren} 
\affiliation{
    \institution{Xi'an Jiaotong University}
}    
\email{rentc2003@stu.xjtu.edu.cn}

\author{Chun Jason Xue} 
\affiliation{
    \institution{Mohamed bin Zayed University of Artificial Intelligence}
}    
\email{jason.xue@mbzuai.ac.ae}

\author{Jen-Ming Wu} 
\affiliation{
    \institution{Hon Hai Research Institute}
}    
\email{jen-ming.wu@foxconn.com}

\author{Nan Guan}
\authornote{Corresponding author.}
\affiliation{
    \institution{City University of Hong Kong}
}    
\email{nanguan@cityu.edu.hk}

% \affiliation{%

% \affiliation{%

% \affiliation{%

% \affiliation{%

% \affiliation{%

% \affiliation{%

% \affiliation{%

%% \renewcommand{\shortauthors}{Trovato et al.}

% \input{sections/0-abstract}
\begin{abstract}
Existing Autonomous Driving Systems (ADS) independently make driving decisions, but they face two significant limitations. First, in complex scenarios, ADS may misinterpret the environment and make inappropriate driving decisions. Second, these systems are unable to incorporate human driving preferences in their decision-making processes.
This paper proposes \emph{Autoware.Flex}, a novel ADS system that incorporates human input into the driving process, allowing users to guide the ADS in making more appropriate decisions and ensuring their preferences are satisfied.
Achieving this needs to address two key challenges: (1) translating human instructions, expressed in natural language, into a format the ADS can understand, and (2) ensuring these instructions are executed safely and consistently within the ADS’ decision-making framework.
For the first challenge, we employ a Large Language Model (LLM) assisted by an ADS-specialized knowledge base to enhance domain-specific translation.
For the second challenge, we design a validation mechanism to ensure that human instructions result in safe and consistent driving behavior.
Experiments conducted on both simulators and a real-world autonomous vehicle demonstrate that \emph{Autoware.Flex} effectively interprets human instructions and executes them safely.
\end{abstract}

% Optional: Comment out the CCS concepts and keywords.
% <concept_id>10010520</concept_id>
% <concept_desc>Computer systems organization</concept_desc>
% <concept_id>10010520.10010570.10010574</concept_id>
% <concept_desc>Computer systems organization~Real-time system architecture</concept_desc>




\maketitle

% \input{sections/1-introduction}
% %\input{sections/2-background}
\section{INTRODUCTION}
Existing Autonomous Driving Systems (ADS) independently make driving decisions based on their perception of the environment~\cite{Autoware2022}~\cite{Apollo2019}. While effective in many scenarios, they still face significant limitations.

\begin{figure}[t]
    \centering
    \includegraphics[width=1\linewidth]{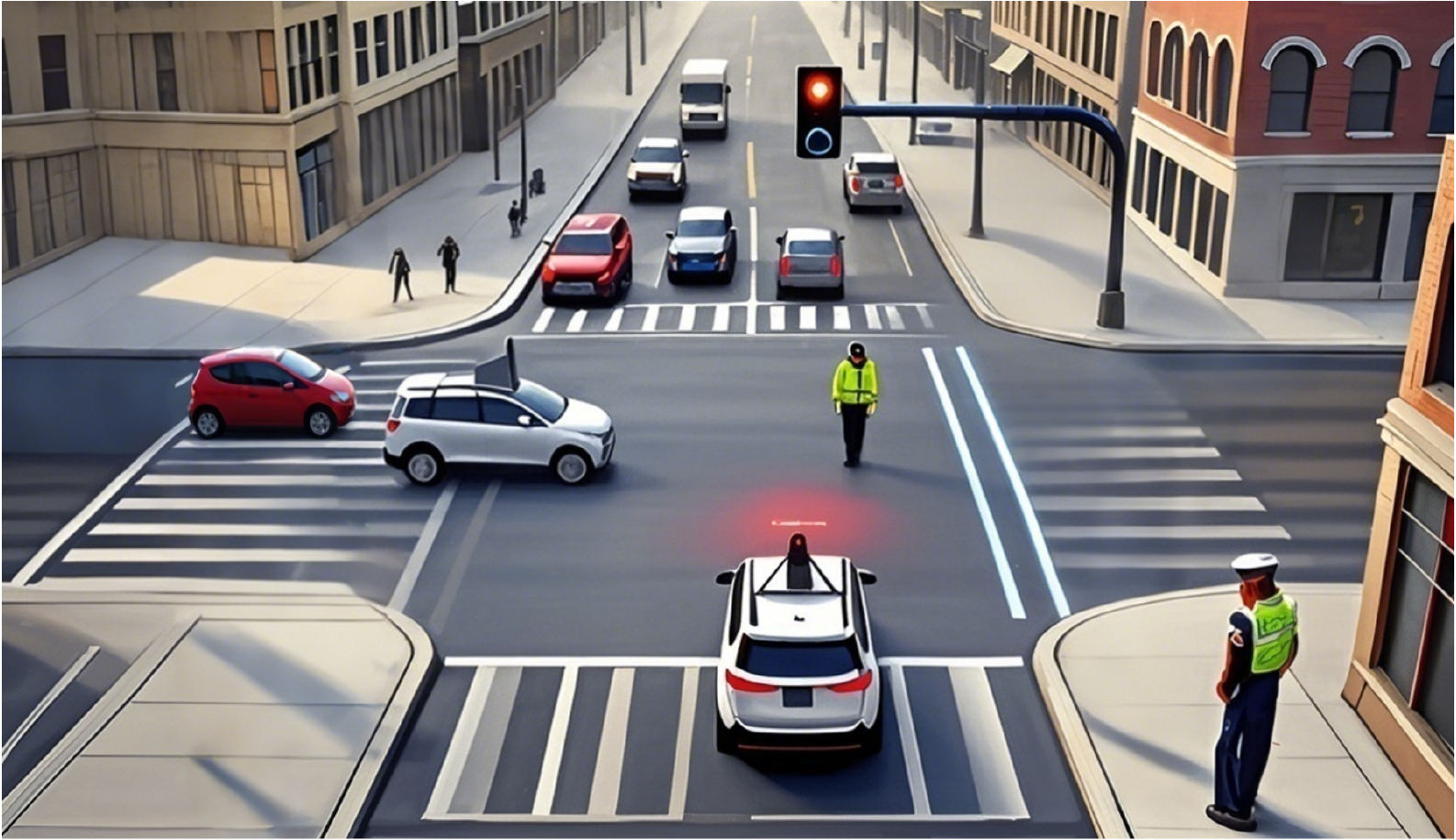}
    \caption{A complex scenario: traffic lights malfunction, and a traffic officer directs vehicles at the intersection.}
    \label{fig:traffic}
    \vspace{-0.1in}
\end{figure}

First, ADS may misinterpret the environment, leading to inappropriate driving decisions in complex scenarios~\cite{mcgregor2018incident}~\cite{khemka2021incident} \\ ~\cite{atherton2022incident}.
For example, consider a road intersection where the traffic lights malfunction and remain stuck on a red signal, as shown in Fig.~\ref{fig:traffic}. To manage traffic, a police officer temporarily directs vehicles at the intersection. While an ADS might be trained to recognize traffic lights and human figures, it could fail to interpret this special situation. Consequently, the ADS might stop the vehicle and wait for the traffic light to turn green. In contrast, a human driver can easily understand the context and follow the instructions of the traffic officer to cross the intersection.

Second, existing ADS do not consider accommodating user-specific driving preferences~\cite{deruyttere2019talk2car}~\cite{hewitt2019assessing}. For example, an ADS typically changes lanes and adjusts the vehicle's speed to optimize traffic flow and avoid blockages. However, a user in an autonomous vehicle might prefer to cruise in the outermost lane at a very low speed while searching for the destination on the roadside. In such cases, the ADS, unaware of the user’s specific requirements, may drive the vehicle in a manner that is safe but inconsistent with the user’s preferences, which can significantly diminish user experience.%which can/which will

To address these limitations, we propose \textit{a novel approach that incorporates human input into ADS's decision-making process}. This approach allows users to guide the ADS through complex scenarios, ensuring more appropriate decisions while also satisfying their personal driving preferences.

Achieving this goal presents two key challenges. The first challenge is \textit{translating human instructions, typically expressed in natural language, into a format that ADS can understand}. While natural language is intuitive for users, ADS systems rely on predefined, structured formats to express information specific to autonomous driving tasks~\cite{pan2024vlp}. A Large Language Model (LLM) could be used for translation~\cite{wang2024grammar}~\cite{shen2024tag}; however, LLMs often lack the domain-specific knowledge required for ADS. To address this, we develop an ADS-specialized knowledge base to provide the LLM with necessary domain-specific information, enabling effective translation. %To address this, we developed an ADS-specialized knowledge base to provide the LLM with necessary domain-specific information, enabling effective translation.

The second challenge is \textit{ensuring that user instructions are executed safely and consistently within the ADS's original decision-making framework}. User instructions cannot always be assumed to be safe. For example, a user might inadvertently issue a command that leads to unsafe driving behavior, such as requesting a lane change while the vehicle is cruising at a high speed. To address this, we develop a mechanism to validate and safeguard user instructions, ensuring they are only executed when safe driving can be guaranteed. This mechanism resolves potential conflicts between user instructions and the ADS's decisions. % To address this, we developed a mechanism to validate and safeguard user instructions, ensuring they are only executed when safe drving can be guaranteed. This mechanism resolves potential conflicts between user instructions and the ADS's decisions.

To implement the proposed approach, we develop Autoware.Flex, a new ADS system built on Autoware.Universe~\cite{AutowareUniverse2022}, the world's leading open-source software for autonomous driving. Experiments are conducted on both a simulation platform (AWSIM)~\cite{AWSIM2023} and a real-world autonomous vehicle prototyped by our team. The results demonstrate that Autoware.Flex effectively interprets and safely executes user instructions, significantly enhancing the capabilities of existing ADS. %To implement the proposed approach, we developed Autoware.Flex, a new ADS system built on Autoware.Universe~\cite{AutowareUniverse2022}, the world's leading open-source software for autonomous driving. Experiments were conducted on both a simulation platform (AWSIM)~\cite{AWSIM2023} and a real-world autonomous vehicle prototyped by our team.

Additionally, we develop an ADS knowledge base to assist domain-specific language translation by LLMs via the Retrieval-Augmented Generation (RAG) architecture~\cite{lewis2020retrieval}. Our knowledge base extracts the key information relevant to autonomous driving decision-making. Experimental results show that our knowledge base achieves higher accuracy in assisting LLMs than standard domain-specific resources, such as the Autoware manual. %we developed an ADS knowledge base
We also create a dataset of ground truths, mapping natural language user instructions to corresponding ADS representations. These resources are valuable for advancing research in this area.

% \input{sections/4-system-design}
% --- LMS ---

% 3.1部分的问题：
% 如前所述，我感觉就只有2个挑战，现在的C1和C2貌似是一个；
% 对语言翻译的挑战，剖析的不够透彻；前面也有这个问题
% C3的问题是，含义不清，目标不具体

% 3.2：等3.1改好了，这段自然跟随过来改好

% 3.3部分的问题：

\section{Autoware.Flex Overview}
\label{s:overview}

\begin{figure}[t]
\centerline{\includegraphics[scale=0.8]{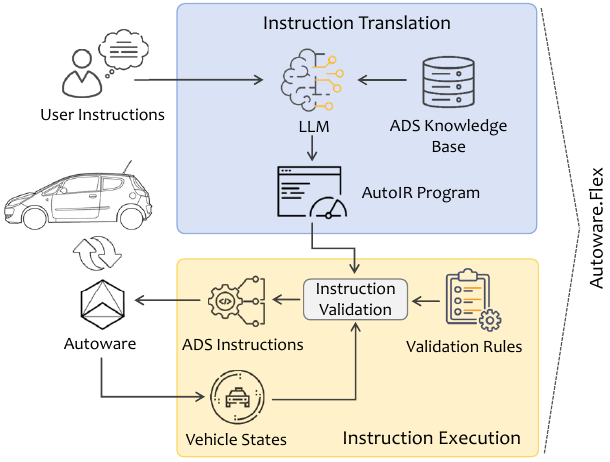}}
\caption{An overview of Autoware.Flex}
\label{f:Overview}
\vspace{-0.1in}
\end{figure}

Autoware.Flex introduces a novel approach to incorporate human input into the decision-making process of an ADS. This allows users to guide the ADS through complex scenarios, enabling more appropriate decisions while also accommodating their personal driving preferences. The architecture of Autoware.Flex is shown in Fig.~\ref{f:Overview}. Autoware.Flex comprises two primary components: Instruction Translation and Instruction Execution, each addressing a key challenge outlined in the introduction.

\textbf{\textit{Instruction Translation.}}

The Instruction Translation component processes user instructions provided in natural language and leverages a Large Language Model (LLM) to generate an AutoIR program — a representation that the ADS can understand. An AutoIR program specifies where in the driving loop the user instruction is injected, as well as key parameters.

While LLMs are adept at understanding human language\\ ~\cite{karanikolas2023large}  \cite{shen2024tag}~\cite{mann2020language}, they typically lack domain-specific knowledge of ADS that is essential for generating accurate AutoIR programs. To address this limitation, we construct an ADS-specific knowledge base that assists the LLM via the Retrieval-Augmented Generation (RAG) architecture. This knowledge base provides the domain knowledge required for the LLM to effectively translate natural language instructions into AutoIR representations. The technical details of this component will be provided in Sec.~\ref{s:trans}. %To address this limitation, we constructed an ADS-specific knowledge base that assists the LLM via the Retrieval-Augmented Generation (RAG) architecture.

\textbf{\textit{Instruction Execution.}}

The Instruction Execution component takes AutoIR programs produced by the Instruction Translation module as input and converts them into actionable ADS instructions. These instructions are then injected into the ADS (in this paper, Autoware.Universe) to influence the vehicle’s behavior.

A critical function of this component is ensuring that user instructions are executed without compromising safety. This is achieved through a rule-based validation process that evaluates the current vehicle status and the environment. The validation checks whether predefined safety rules — derived from human expertise on how ADS parameters affect driving behavior — are satisfied. If the validation succeeds, the AutoIR program is translated into ADS instructions. This translation step is straightforward and ensures seamless integration with the ADS. The validated instructions are then sent to the ADS for execution, enabling safe and user-guided driving behaviors. The technical details of this component will be provided in Sec.~\ref{s:execute}.

% ==================

\section{Translating User Instructions}
\label{s:trans}

\begin{figure*}[t]
\centerline{\includegraphics[scale=0.74]{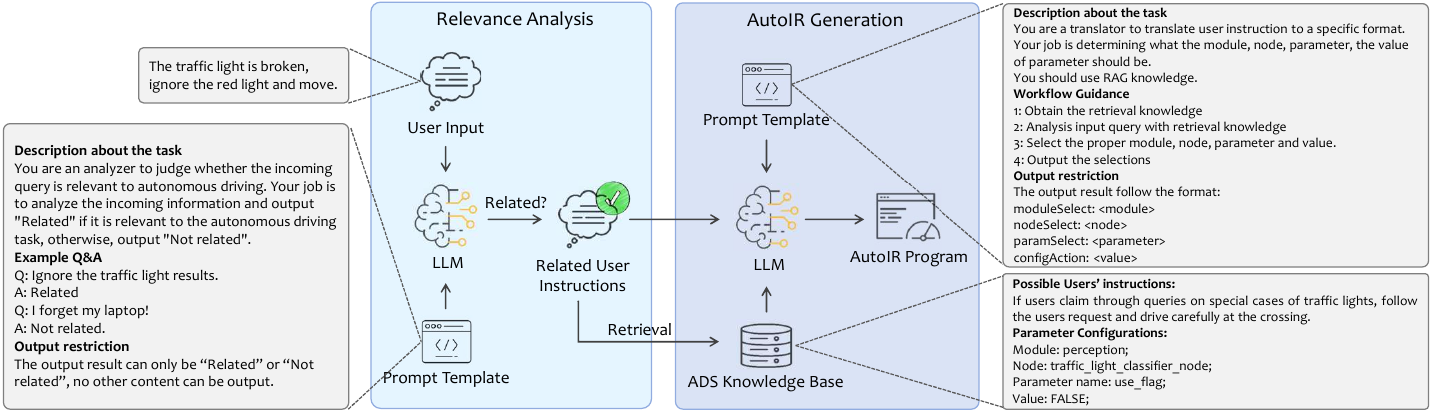}}
\caption{The workflow of user instruction translation}
\label{f:Translating Instructions}
\end{figure*}

Translating user instructions into AutoIR programs involves two main steps. First, we determine whether a user instruction is relevant to autonomous driving. Only instructions related to driving are processed further. Second, if the user instruction is relevant, it serves as the input to generate the corresponding AutoIR program. The detailed workflow is shown in Fig.~\ref{f:Translating Instructions}.

\subsection{Relevance Analysis}
\label{ss:relevance}

When a user is in an autonomous vehicle, their conversations may cover a wide range of topics, many of which might not be related to autonomous driving. To ensure the system processes only relevant input, we need to filter out unrelated dialogue, focusing solely on instructions pertaining to the vehicle's operation.

To achieve this, we leverage the capabilities of LLMs to perform this classification task. Instead of re-training the LLM, we adopt in-context learning, specifically Chain of Thought (CoT) prompting~\cite{wei2022chain}, a technique that improves accuracy by providing context-specific examples within the prompt. Specifically, we feed the LLM not just the user’s input but also a carefully designed prompt template. This template includes descriptive information about the task and several Q$\&$A examples illustrating how to identify autonomous driving-related instructions (The left side of Fig.~\ref{f:Translating Instructions} gives examples of user input and prompt template).

If the LLM determines that the input sentence qualifies as a user instruction, it forwards the instruction to the next step for further processing. This approach ensures a lightweight yet effective method for filtering user input without requiring extensive model customization.

\subsection{AutoIR Generation}
\label{ss:autoirgen}

In the second step, we generate an AutoIR program to implement a user instruction. Before diving into the details, we briefly introduce AutoIR. AutoIR is a custom-designed language that standardizes the translation output into a format that is understandable by the ADS. Essentially, it maps user instructions into Autoware's software constructs. Since Autoware is built on the ROS 2 middleware, we will first provide an overview of the structure of Autoware and ROS 2. This foundation will help readers to understand the role and design of AutoIR within the system.

\begin{figure}[t]
\centerline{\includegraphics[scale=0.4]{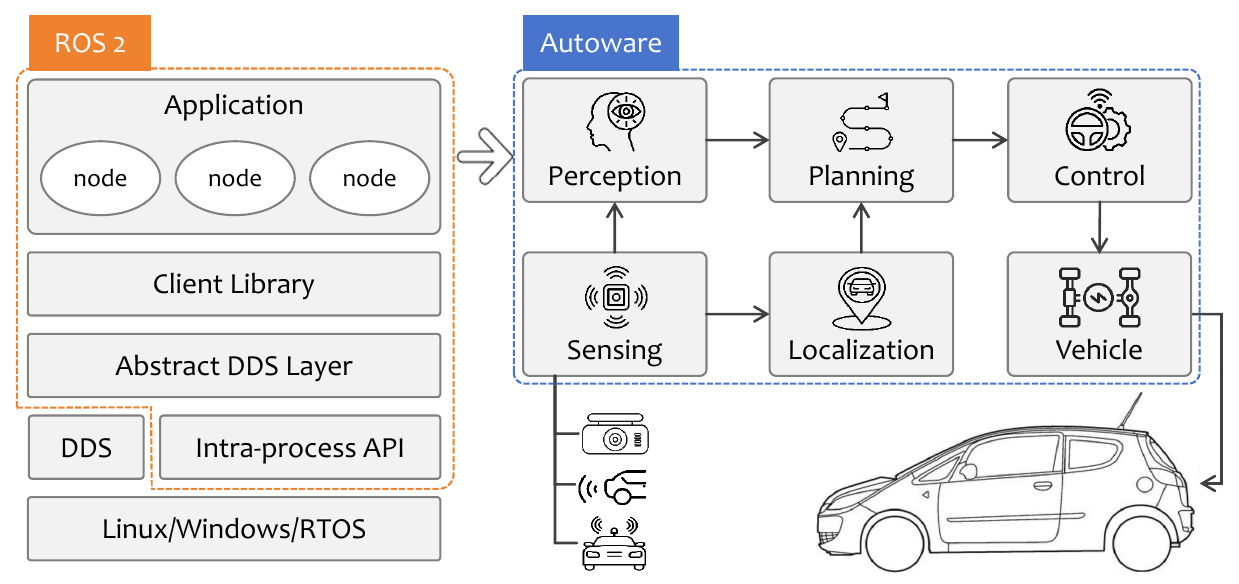}}
\caption{The architectures of Autoware and ROS 2}
\label{f:autowareros2}
\vspace{-0.1in}
\end{figure}

\vspace{0.05in}
\subsubsection{The architecture of Autoware and ROS 2}

\vspace{0.05in}

Autoware is an open-source software framework specifically designed to address the complexities of autonomous driving systems~\cite{Autoware2022}. It employs a modular architecture that integrates all critical components required for autonomous vehicle operation, including sensing, localization, perception, planning, and control (as illustrated in Fig.~\ref{f:autowareros2}). 
The sensing module collects raw environmental data from various sensors, such as LiDAR, cameras, and radar. This data is processed by the localization module, which determines the precise position and orientation of the vehicle within its environment. The perception module interprets the sensor data to identify objects, detect obstacles, and understand the surrounding environment. Based on this information, the planning module develops driving strategies, routes, and trajectories tailored to the vehicle's goals and the environmental context. Finally, the control module executes these planned actions by managing the vehicle’s actuators, such as steering, acceleration, and braking. These actions are further converted into low-level driver steps by the Vehicle module.
These modules collaborate to enable efficient and accurate decision-making processes for autonomous vehicles.

ROS 2 (Robot Operating System 2) is an open-source framework for developing modular, scalable, and secure robotic software~\cite{macenski2022ros2}. It is widely used in diverse applications, including autonomous driving, multi-robot systems, industrial automation, and healthcare robotics.

In ROS 2, an application is built around nodes, which are lightweight, modular components that communicate using a publish-subscribe model (DDS in Fig.~\ref{f:autowareros2}). Each Autoware module comprises a collection of ROS 2 nodes, each dedicated to specific tasks such as processing sensor data, detecting obstacles, or generating driving trajectories. This node-based design enables distributed processing and provides fine-grained control over the vehicle's functionalities.
Key layers of ROS 2, including the Client Library, Abstract DDS Layer, DDS, and Intra-process API, offer the necessary abstractions for efficient inter-node communication and seamless data exchange. By leveraging ROS 2, Autoware ensures robust communication between nodes, enabling reliable handling of tasks ranging from sensor data ingestion to high-level decision-making and vehicle control.

\begin{figure}[t]
\centerline{\includegraphics[scale=0.9]{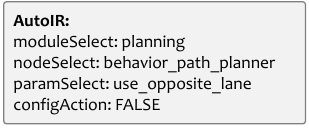}}
\caption{An example of an AutoIR program}
\label{f:autoir}
\vspace{-0.1in}
\end{figure}

\vspace{0.05in}
\subsubsection{AutoIR Semantics and Generation}

\vspace{0.05in}

The semantics of AutoIR define how user instructions are translated into metadata used for ROS 2 implementation. An AutoIR program consists of several domains of information, as exemplified in Fig.~\ref{f:autoir}.
The \texttt{moduleSelect} domain specifies the Autoware module that the user instruction will impact. For instance, if the user requests the ADS to change lanes, this instruction will affect the planning module. The \texttt{nodeSelect} domain identifies the specific node within the selected module that will be influenced, as each Autoware module may consist of multiple nodes. The \texttt{paramSelect} domain provides parameters for the selected node, guiding it to execute the desired actions. The \texttt{configAction} domain specifies the values to be assigned to these parameters, ensuring the node performs actions accordingly. The \texttt{Timer} domain specifies the lifetime of the user instruction.

To translate natural language user instructions into AutoIR, we leverage a Large Language Model (LLM). However, a major challenge lies in the fact that LLMs typically lack domain-specific knowledge of ADS and are unfamiliar with AutoIR semantics~\cite{fan2024survey}. While one possible solution is to re-train the LLM with AutoIR examples and ADS domain knowledge, this approach is resource-intensive and requires a large amount of training data. Instead, we adopt the Retrieval-Augmented Generation (RAG) approach, which equips the LLM with an external knowledge base. RAG allows the LLM to retrieve relevant information during task execution, significantly reducing the need for re-training.

The effectiveness of RAG heavily depends on the quality of the knowledge base~\cite{fatehkia2024t}. For example, one can use the entire Autoware manual as the knowledge base, but this is ineffective, as the manual contains a large amount of unrelated information, making it difficult for the LLM to locate the specific details it needs. To address this, we build a specialized ADS knowledge base derived from Autoware documentation. Each entry in the knowledge base pairs a driving scenario (representing a type of user instructions) with the corresponding AutoIR program. An example is illustrated in Fig.~\ref{f:Translating Instructions}.
%To address this, we built a specialized ADS knowledge base derived from Autoware documentation. Each entry in the knowledge base pairs a driving scenario (representing a type of user instructions) with the corresponding AutoIR program. An example is illustrated in Fig.~\ref{f:Translating Instructions}.

During AutoIR generation, the user instruction serves as input, triggering a retrieval query on the ADS knowledge base to extract relevant information. The retrieved information, along with the original user instruction, is then fed to the LLM to generate the corresponding AutoIR program. To further improve accuracy, we provide a structured prompt template as part of the input. This template guides the LLM on how to utilize the retrieved knowledge effectively during the generation process.

\section{Executing User Instructions}
\label{s:execute}

The execution of user instructions involves dynamically reconfiguring the ADS using the detailed information contained in AutoIR programs, enabling the ADS execution loop to carry out the specified driving actions.

A key challenge in this process is ensuring that user instructions do not result in unsafe driving behaviors. User instructions are often issued in special-case scenarios where the user's intent may conflict with the ADS's predefined rules. For instance, consider the malfunctioning traffic light scenario depicted in Fig.~\ref{fig:traffic}. In this situation, the user instructs the ADS to proceed through the intersection, which directly contradicts the ADS’s predefined rule: “When a red light is observed at an intersection, keep the vehicle stationary.” In these scenarios, following the user’s instruction places the responsibility for safety on the user, as their command overrides the system’s default behavior.
Even though, the system should try to avoid executing instructions that are intentionally or carelessly unsafe. For instance, if the user instructs the vehicle to change lanes while cruising at high speed, such an instruction can be blocked to prevent potential accidents and preserve safety.

To address this, we develop a rule-based mechanism to validate and safeguard user instructions before integrating them into the ADS system. This mechanism ensures that user instructions meet predefined safety criteria and reduces the risks associated with unsafe commands.
% To address this, we developed a rule-based mechanism to validate and safeguard user instructions before integrating them into the ADS system. This mechanism ensures that user instructions meet predefined safety criteria and reduces the risks associated with unsafe commands.

\begin{figure*}
    \centering
    \includegraphics[width=0.95\linewidth]{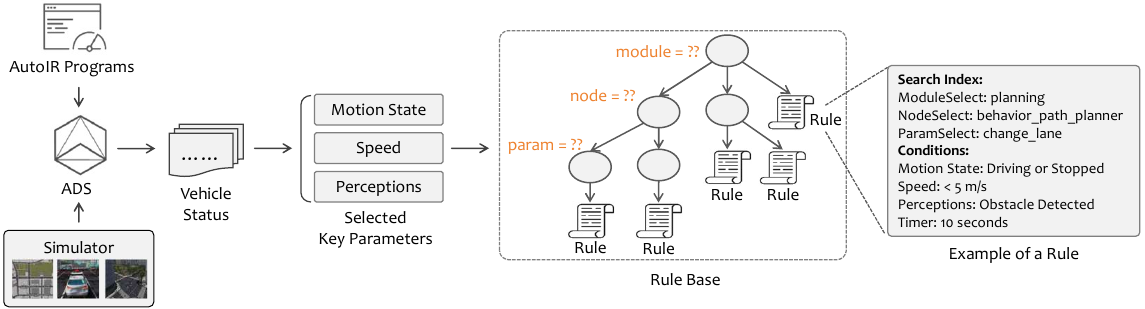}
    \caption{The workflow for the design of the rule base}
    \label{f:rulebase}
\end{figure*}

\subsection{Rule Base Design}
\label{ss:rulebase}

We design the rules to safeguard user instruction execution offline using a simulation-based approach. This process begins by generating a set of AutoIR programs, which are manually validated for correctness. These programs are then tested in an ADS simulator to replicate typical driving scenarios. For each scenario, an AutoIR program is inserted, and the vehicle’s status is observed through the ADS software. While the ADS provides numerous vehicle status parameters, we focus on the key parameters necessary for defining the rules:

\begin{itemize}
\item \texttt{Motion State}: indicates whether the vehicle is moving or stopped, along with the reasons for stopping.
\item \texttt{Speed}: specifies the vehicle's current speed.
\item \texttt{Perceptions}: provides information about the objects identified by the vehicle that may influence its driving decisions.
\end{itemize}

For example, to enforce a safety requirement such as ``if the speed is above 70 km/h, do not change lanes'', this condition is implemented in the simulated driving scenarios. By reading the key parameters of the vehicle's status during simulation, we derive rules that ensure safe instruction execution.

Each rule consists of two key components: the ``Search Index'' and ``Conditions''. The Search Index is used to identify and retrieve the relevant rule from the rule base based on the information provided by the AutoIR program. The Conditions specify safeguard parameters that ensure safe system behavior during the execution of user instructions.

An important condition is the \texttt{timer}, which serves as a critical safety mechanism. Since user instructions may override the ADS's default rules, the \texttt{timer} defines a specific duration, after which the system reverts to its default settings. This rollback mechanism ensures that deviations from standard behavior are temporary and safety is preserved. Currently, the timer values are manually and conservatively encoded in the rules during their design.

The rules generated through this process are organized into a tree structure to optimize searching. When an AutoIR program is received, the corresponding rule is located in the rule base using the search index, which facilitates efficient navigation and retrieval. Fig.~\ref{f:rulebase} illustrates the workflow for generating the rule base, along with an example of a rule.

Upon receiving a user instruction in the form of an AutoIR program, execution proceeds only if the program matches a rule in the rule base and the specified conditions are satisfied. (The detailed validation process is introduced in the following subsection.) We note that the number of driving scenarios and AutoIR programs used during rule generation limits the number of generated rules and, consequently, the scope of acceptable user instructions.
While it is impossible to enumerate all potential driving scenarios and user instructions due to their unlimited number, our current approach generates rules that reflect typical driving scenarios. But this ensures a safety baseline: unmatched user instructions and those that do not satisfy the conditions are ignored. System designers can, however, incrementally expand the rule base to accommodate more driving scenarios and support a broader range of user instructions over time.

\begin{figure}
    \centering
    \includegraphics[width=0.85\linewidth]{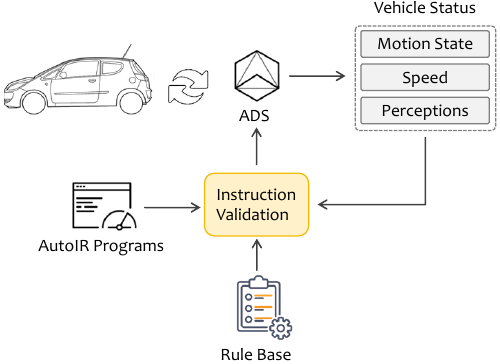}
    \caption{The workflow of instruction validation}
    \label{f:validation}
\end{figure}

\subsection{Runtime Instruction Validation}
\label{ss:validation}

At runtime, a dedicated software component is responsible for validating user instructions. The validation workflow is illustrated in Fig.~\ref{f:validation}. Importantly, whether a user instruction passes validation depends on the vehicle's status, which continuously changes during driving. As a result, upon the arrival of a user instruction, the validation process must repeatedly evaluate the instruction until its lifetime expires. Currently, the lifetime of a user instruction is manually set to 10 seconds. This setting can, of course, be further optimized based on the specific requirements of different instructions.

A user instruction represented as an AutoIR program is matched against the rules in the rule base, along with real-time vehicle status data retrieved from the ADS. Only instructions that successfully pass validation, ensuring safety, are issued to the ADS for execution.

It is worth noting that each AutoIR program undergoes a final transformation into low-level ADS instructions before being fed to the ADS. This transformation is straightforward and does not require further elaboration here.

The instruction validation algorithm is presented in Algorithm 1. This algorithm takes user instructions in AutoIR format ($I$) and the rule base ($R$) as input and outputs whether the given user instruction should be executed (via an activation signal). Based on the information in the AutoIR program, Line 2 performs a search to find the corresponding rule in the rule base. Lines 3–8 describe the runtime checking process, during which the vehicle’s status is continuously retrieved from the ADS to determine whether the conditions are met to execute the user instruction. This process terminates when the user instruction expires.% This process terminates if the user instruction is activated or expires.

\begin{algorithm}[t]
\begin{algorithmic}[1]
{\small
	%\Statex \textbf{Input} The current AutoIR $AutoIR$, the rule base $Rules$
    \REQUIRE Input AutoIR $I$, Rule Base $R$
	% \Statex \textbf{Output} Activating Signal $Activate$
    \ENSURE Activation Signal
	\STATE \textbf{Function} Instruction\_Validation $(I, R)$
    \STATE \ \ \ \ \ \ \ $Rule \leftarrow Rule\_Searching (I,R)$
    \STATE \ \ \ \ \ \ \ \textbf{while}\ ($Instruct~not~expired$) \textbf{do}
    \STATE \ \ \ \ \ \ \ \ \ \ \ \ \ \ \ $cur\_status\leftarrow Current\ vehicle\ status$
    \STATE \ \ \ \ \ \ \ \ \ \ \ \ \ \ \ \textbf{if} $Matching(cur\_status,\ Rule) = TRUE $ \textbf{then}
    \STATE \ \ \ \ \ \ \ \ \ \ \ \ \ \ \ \ \ \ \ \ \ \ \ \ \ \ \ \ \textbf{return} $Activated$
    \STATE  \ \ \ \ \ \ \ \ \ \ \ \ \ \ \ \textbf{end if}
    \STATE \ \ \ \ \ \ \ \textbf{end while}
    \STATE \textbf{return} $Not\_activated$
    \STATE \textbf{end Function}   
}
\end{algorithmic}
\caption{Instruction Validation Algorithm (IVA)}
\label{algo:condition checking}
\end{algorithm}

\section{IMPLEMENTATION}

We implement the proposed Autoware.Flex system based on Autoware.Universe~\cite{AutowareUniverse2022}. Autoware.Universe is installed on Ubuntu 22.04 with ROS 2 Humble~\cite{humble} serving as the middleware.
% We implemented the proposed Autoware.Flex system based on Autoware.Universe~\cite{AutowareUniverse2022}. Autoware.Universe is installed on Ubuntu 22.04 with ROS 2 Humble~\cite{humble} serving as the middleware.

\textbf{\textit{Instruction Translation based on LLM}}

For instruction translation, we utilize QWenVL~\cite{bai2023qwen}, specifically the QWenVL-Max version, a state-of-the-art Large Language Model (LLM) developed by Aliyun. The QWenVL services are accessed via the Dashscope library~\cite{dashscope}, while the LangChain framework~\cite{langchain} is employed to implement the Retrieval-Augmented Generation (RAG) framework. LangChain also provides standard embedding tools to facilitate this implementation.
Our ADS knowledge base is segmented into chunks, each containing 700 tokens. These chunks are converted into vectors and stored in a FAISS vector database~\cite{douze2024faiss}, enabling efficient retrieval and processing as part of the RAG framework.

\textbf{\textit{Integration with Autoware on ROS 2}}

Since Autoware.Flex is built on Autoware.Universe, which itself is implemented using ROS 2, the two primary components of our system, instruction translation and instruction execution, are also implemented as ROS 2 nodes. These nodes are seamlessly integrated into the existing Autoware architecture.

For communication between ROS 2 nodes, we rely on ROS 2's publish-subscribe asynchronous communication model. In this model, nodes exchange data indirectly via topics, which act as data channels. Each topic is associated with a specific message type that defines the structure of the data being transmitted. Nodes can publish messages to a topic or subscribe to receive messages from a topic to facilitate data exchange.

We define two new topics for the Autoware.Flex nodes. Topic \texttt{user\_instruction} serves as the input for user instructions in natural language, directed to the instruction translation node.
Topic \texttt{AutoIR} is used to transfer the AutoIR programs generated by the instruction translation node to the instruction execution node.
% We defined two new topics for the Autoware.Flex nodes. Topic \texttt{user\_instruction} serves as the input for user instructions in natural language, directed to the instruction translation node.

\textbf{\textit{Vehicle Status and Command Execution}}

The instruction execution node also interacts with the ADS by reading vehicle status data and issuing commands to control autonomous driving behaviors. To access vehicle status data, the instruction execution node subscribes to the following topics already established in the Autoware implementation:

{\small \begin{itemize} \item \texttt{/api/motion/state}
\item \texttt{/vehicle/status/velocity\_status}
\item \texttt{/perception/object\_recognition/detection/objects}
\item \texttt{/perception/traffic\_light\_recognition/traffic\_li\\ght/detection/rois}
\end{itemize} }

Once a user instruction is validated and activated, the corresponding command is issued to the relevant Autoware modules using the \texttt{param} mechanism. The command format is as follows:

{\small \texttt{ros2 param set <module> <node> <param> <config\_action>} } 
\indent An important detail is that when Autoware.Flex executes this command, and the original parameters of the related ROS 2 nodes are temporarily overridden. To ensure system integrity, our system first backs up the original parameters and automatically restores them once the timer for the user instruction expires.

\begin{table*}
\scalebox{0.75}{
\renewcommand{\arraystretch}{1.5} % for larger row height
\begin{tabular}{|cccccc|}
\hline
\multicolumn{1}{|c|}{Methods}             & \multicolumn{1}{c|}{ModuleSelect Accuracy (\%)} & \multicolumn{1}{c|}{NodeSelect Accuracy (\%)} & \multicolumn{1}{c|}{ParamSelect Accuracy (\%)} & \multicolumn{1}{c|}{ConfigAction Accuracy (\%)} & Overall Accuracy (\%) \\ \hline
\multicolumn{1}{|c|}{Our Knowledge Base} & \multicolumn{1}{c|}{95.5}                       & \multicolumn{1}{c|}{95.5}                     & \multicolumn{1}{c|}{93.5}                      & \multicolumn{1}{c|}{87}                         & 87                    \\ \hline
\multicolumn{1}{|c|}{Autoware Manual}           & \multicolumn{1}{c|}{81}                         & \multicolumn{1}{c|}{64.5}                     & \multicolumn{1}{c|}{50}                        & \multicolumn{1}{c|}{32}                         & 32                    \\ \hline
\end{tabular}}
\caption{The results of accuracy evaluation for user instruction translation}
\label{t:Accuracy}
\vspace{-0.1in}
\end{table*}

\section{Experiments and Evaluation}
\label{s:eval}

In this section, we intend to evaluate (1) the accuracy and latency performance of instruction translation (Sec.~\ref{ss:eval_trans}), (2) the effectiveness of instruction execution based on a simulation platform (Sec.~\ref{ss:eval_exe}), and (3) the effectiveness of the overall Autoware.Flex system on a real-world autonomous vehicle (Sec.~\ref{ss:eval_auto}).

\subsection{Evaluation of Instruction Translation}
\label{ss:eval_trans}

\vspace{0.05in}
\subsubsection{The AutoIR Dataset}

\vspace{0.05in}

To evaluate the accuracy of user instruction translation, ground truths are essential. To this end, we develop a custom AutoIR dataset based on an in-depth analysis of Autoware to serve as the ground truth. This dataset is specifically created to address the absence of benchmarks in the existing literature for translating natural language into AutoIR.
% To evaluate the accuracy of user instruction translation, ground truths are essential. To this end, we developed a custom AutoIR dataset based on an in-depth analysis of Autoware to serve as the ground truth. This dataset was specifically created to address the absence of benchmarks in the existing literature for translating natural language into AutoIR.
The dataset comprises pairs of user instructions in natural language and their corresponding AutoIR programs. These AutoIR programs are carefully crafted based on our extensive experience with Autoware and further verified through simulation to ensure they result in the correct driving behavior. The dataset includes 170 such pairs. Additionally, we incorporate 30 natural language user instructions unrelated to autonomous driving intended for testing purposes.
% The dataset comprises pairs of user instructions in natural language and their corresponding AutoIR programs. These AutoIR programs were carefully crafted based on our extensive experience with Autoware and further verified through simulation to ensure they result in the correct driving behavior. The dataset includes 170 such pairs. Additionally, we incorporated 30 natural language user instructions unrelated to autonomous driving intended for testing purposes.

\vspace{0.05in}
\subsubsection{Accuracy Evaluation}
~\par
\vspace{0.05in}

\textbf{\textit{(1) Evaluation targets}}

We aim to evaluate not only the overall user instruction translation function but also its individual components, including relevance analysis, module selection, node selection, parameter selection, and configuration value assignment.

\textbf{\textit{(2) Compared approaches}}

\textit{- Relevance Analysis Evaluation}

\begin{figure}[t]
    \centering
    \includegraphics[width=1\linewidth]{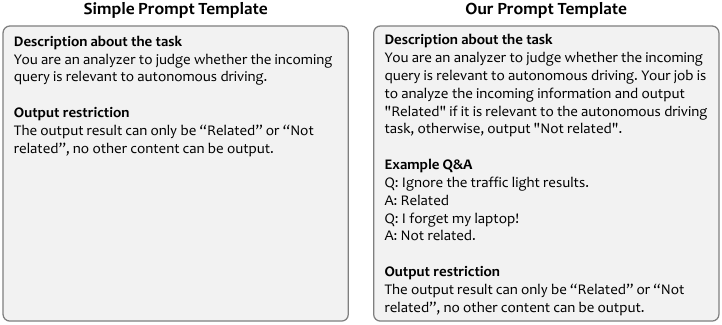}
    \caption{Examples of the Simple Prompt template and our proposed prompt template for relevance analysis}
    \label{f:prompts}
\end{figure}

To assess the relevance analysis component, we focus on evaluating the effectiveness of our proposed prompt template, which leverages in-context learning. For comparison, we introduce a baseline template called \texttt{Simple Prompt}. The \texttt{Simple Prompt} template contains only a description of the relevance analysis task and output format constraints without providing Q\&A examples. In contrast, our prompt template includes Q\&A examples to guide the LLM more effectively. Examples of both templates are shown in Fig.~\ref{f:prompts}.

\textit{- Evaluation of Other Components}

To evaluate the other components and the overall user instruction translation function, the knowledge base that assists the LLM plays a critical role. In these experiments, we compare our specialized ADS knowledge base with a baseline knowledge base directly using the Autoware user manual. For consistency, our proposed prompt template is used during the relevance analysis step in all evaluations.

\textbf{\textit{(3) Evaluation metric}}

For all components and the overall translation function, accuracy is defined as the ratio of correctly processed results to the total number of test cases (200 items from the AutoIR dataset).

\textbf{\textit{(4) Accuracy evaluation results}}

In terms of relevance analysis accuracy, our approach achieves $99\%$, significantly outperforming the \texttt{Simple Prompt}, which achieves only $92\%$. This demonstrates that the in-context learning prompting approach is more effective in guiding the LLM to correctly determine whether a user instruction in natural language is relevant.

The accuracy results for other components are summarized in Table~\ref{t:Accuracy}. Our specialized knowledge base shows a substantial improvement in accuracy compared to the approach using the Autoware manual as the knowledge base. This improvement is observed not only in the overall accuracy but also across all individual components.
Notably, the ConfigAction step has the most significant impact on the overall accuracy of user instruction translation. This finding provides valuable insights for future optimization of the knowledge base, emphasizing the importance of enhancing the accuracy of the ConfigAction step.

\vspace{0.05in}
\subsubsection{Latency Evaluation}

\vspace{0.05in}

\textbf{\textit{(1) Evaluation targets \& compared approaches}}

We evaluate the execution time overhead for three components: the relevance analysis step, the translation step, and the end-to-end user instruction translation process (including the latency of the first two steps).

For relevance analysis, we compare the latency of our approach with the \texttt{Simple Prompt} approach. For the translation step, we compare our approach with an alternative approach that uses the Autoware manual as the knowledge base in the RAG framework. To evaluate end-to-end latency, we consider four configurations combining the approaches used in the relevance analysis and translation steps.

\textbf{\textit{(2) Evaluation metric}}

We measure the time spent on each evaluation target (in units of seconds). The latency results were obtained by executing the systems on a desktop computer equipped with an Intel Core i7-10700 CPU running at 2.90 GHz.
% We measured the time spent on each evaluation target (in units of seconds). The latency results were obtained by executing the systems on a desktop computer equipped with an Intel(R) Core(TM) i7-10700 CPU running at 2.90 GHz.

\begin{figure*}[t]
\centerline{\includegraphics[scale=0.28]{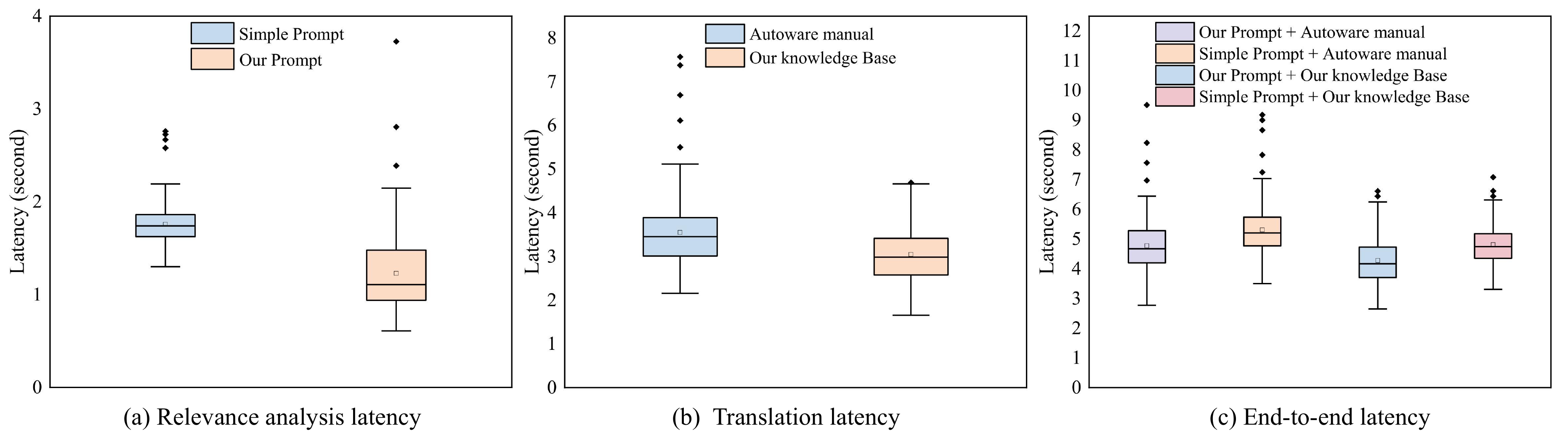}}
\caption{The results of latency evaluation for user instruction translation}
\label{f:Latency}
\end{figure*}

\textbf{\textit{(3) Latency evaluation results}}

The experimental results for latency evaluation are presented in Fig.~\ref{f:Latency}.

\textit{- Relevance Analysis Step}

Fig.~\ref{f:Latency} (a) shows the latency results for the relevance analysis step. Our prompt template outperforms the compared approach in both average latency (1.5 s vs. 1.8 s) and peak latency (2.0 s vs. 2.5 s). This demonstrates that our prompt template not only improves accuracy (as evaluated previously) but also reduces the time overhead on the LLM to determine relevance. Further analysis reveals that latency variations are primarily due to the QWenVL-Max Web API, which is susceptible to network fluctuations.

\textit{- Translation Step}

Fig.~\ref{f:Latency} (b) compares the latency of using different knowledge bases in the translation step. Once again, our approach outperforms the alternative approach that uses the Autoware manual as the knowledge base. This suggests that our specialized ADS knowledge base not only enhances accuracy but also reduces the computational burden on the LLM during the RAG process, thanks to the significantly smaller size of our ADS knowledge base.

\textit{- End-to-End Latency}

Fig.~\ref{f:Latency} (c) illustrates the end-to-end latency results for the four configurations. This latency includes not only the time spent on the two individual steps but also the additional delays introduced by data communication between them. Our system, which employs the in-context learning prompt template for relevance analysis and utilizes our ADS knowledge base in the RAG framework, achieves the best performance, demonstrating the lowest average and peak latency.

\textit{- Overall Analysis}

The observed latency values are within an acceptable range, considering normal time delays in the autonomous driving loop. In emergency situations, such as a sudden lane change to avoid an unexpected obstacle, existing ADS systems should perform better than human intervention and should handle such cases autonomously without user involvement.

\begin{figure*}[t]
\centerline{\includegraphics[scale=0.6]{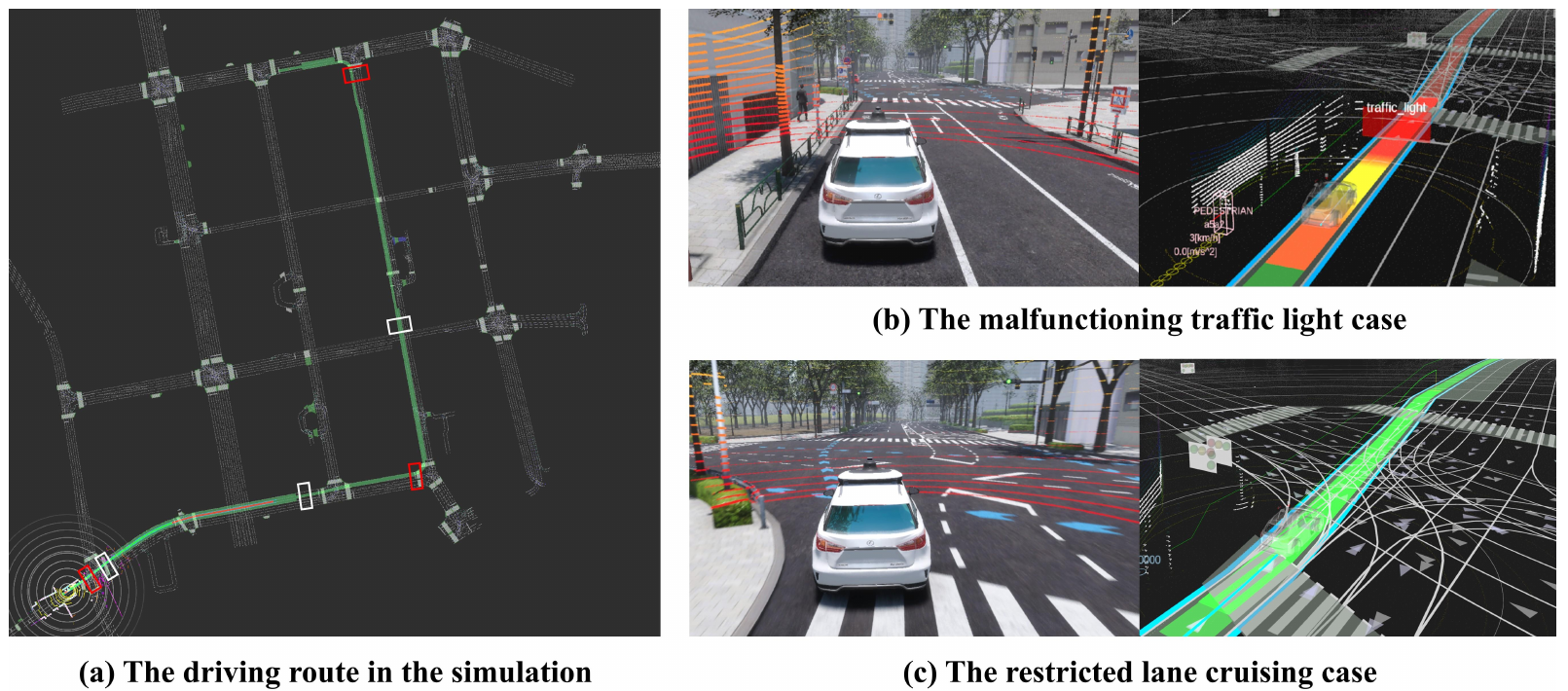}}
\caption{Two scenarios to evaluate user instruction execution. (a) is the HDMap used in AWSIM, i.e., the routes taken by the vehicle. (b) shows the malfunctioning traffic light case; (c) shows the restricted cruising lane case.}
\label{f:twocases}
\end{figure*}

\begin{figure}[t]
\centerline{\includegraphics[scale=0.31]{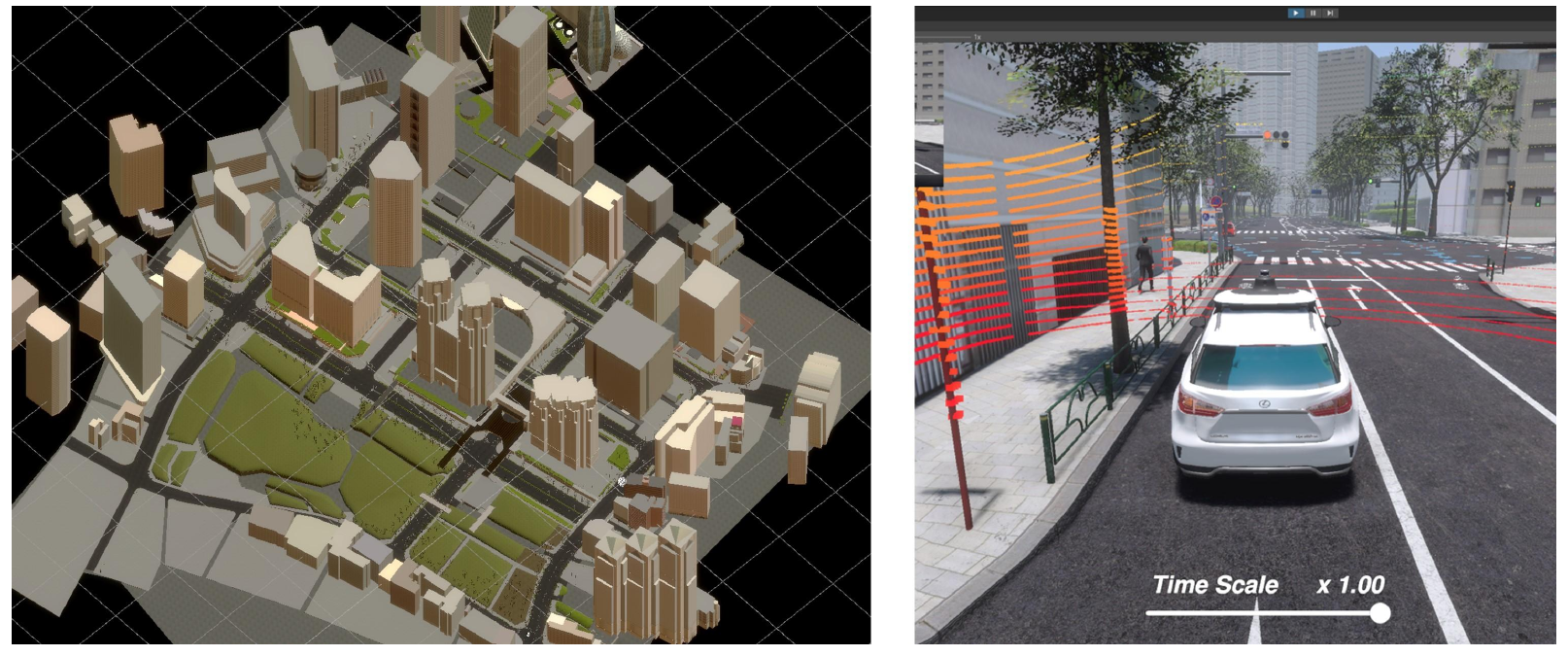}}
\caption{The simulated driving environment in the AWSIM simulator. The left-side picture is an overview of the simulated Shinjuku district in Unity3D; the right-side picture shows the overview of the unity project.}
\label{f:AWSIM}
\end{figure}

\subsection{Evaluation of Instruction Execution}
\label{ss:eval_exe}

To evaluate the effectiveness of Autoware.Flex in executing user driving instructions, we employ the state-of-the-art Autoware simulator, AWSIM~\cite{AWSIM2023}. To simulate complex driving environments, we modified the Unity3D project of AWSIM.
%To evaluate the effectiveness of Autoware.Flex in executing user driving instructions, we employed the state-of-the-art Autoware simulator, AWSIM~\cite{AWSIM2023}. To simulate complex driving environments, we modified the Unity3D project of AWSIM.
The default AWSIM environment is based on the Shinjuku district in Japan and features a highly detailed 3D point cloud environment with semantic maps. Fig.~\ref{f:AWSIM} illustrates the simulated Shinjuku district environment and the Unity3D project setup used for AWSIM.

We design two scenarios within the simulator to validate whether our system can effectively and safely execute user instructions:
% We designed two scenarios within the simulator to validate whether our system can effectively and safely execute user instructions:

\textbf{Malfunctioning traffic light}: In this scenario, the traffic lights at an intersection malfunction, continuously displaying red lights. A traffic officer is assumed to be directing vehicles (though, due to the simulator's limitations, the officer is not visually represented in the scene). The user, understanding the situation, issues an instruction for the vehicle to ignore the traffic light and proceed through the intersection. Figure \ref{f:twocases} (b) illustrates this scenario.

\textbf{Restricted lane cruising}: In this scenario, the user is searching for a destination building along the roadside. To facilitate the search, the user instructs the vehicle to cruise exclusively in the outermost lane (note that the map in the simulator is in Japan; the outermost lane is the left-most lane), making it easier to locate the destination. Figure \ref{f:twocases} (c) depicts this scenario.

For each scenario, we use three different sentences to express the corresponding user instruction. For comparison, we also test native Autoware under the same conditions (without user instructions) to observe how it handled these scenarios. The simulation results are summarized in Table~\ref{tab:sim}.
% For each scenario, we used three different sentences to express the corresponding user instruction. For comparison, we also tested native Autoware under the same conditions (without user instructions) to observe how it handled these scenarios. The simulation results are summarized in Table~\ref{tab:sim}.

In the malfunctioning traffic light scenario, three distinct user instructions are issued with the same intent to instruct the vehicle to proceed. Autoware.Flex successfully interpret and executed all three instructions. The simulation results show that the vehicle correctly move through the intersection by following the user’s instructions. In contrast, native Autoware adhere strictly to its predefined rule: ``I must wait until the traffic light turns green'', and remain stationary. Table~\ref{tab:sim} also lists the generated AutoIR programs and the corresponding rules matched for this scenario.
% In the malfunctioning traffic light scenario, three distinct user instructions were issued with the same intent to instruct the vehicle to proceed. Autoware.Flex successfully interpreted and executed all three instructions. The simulation results showed that the vehicle correctly moved through the intersection by following the user’s instructions. In contrast, native Autoware adhered strictly to its predefined rule: ``I must wait until the traffic light turns green'', and remained stationary. Table~\ref{tab:sim} also lists the generated AutoIR programs and the corresponding rules matched for this scenario.

In the restricted lane cruising scenario, we again issue three different user inputs, all instructing the vehicle to remain in the outermost lane. The simulation results confirm that the vehicle stay in the outermost lane as directed. In comparison, native Autoware is unable to maintain the outermost lane and followed its own lane-selection rules, such as changing lanes based on traffic conditions. This user instruction is also associated with a timer specifying its validity duration. After the timer expired, the ADS revert to its predefined lane-control rules.
% In the restricted lane cruising scenario, we again issued three different user instructions, all instructing the vehicle to remain in the outermost lane. The simulation results confirmed that the vehicle stayed in the outermost lane as directed. In comparison, native Autoware was unable to maintain the outermost lane and followed its own lane-selection rules, such as changing lanes based on traffic conditions. This user instruction was also associated with a timer specifying its validity duration. After the timer expired, the ADS reverted to its predefined lane-control rules.

We also measure the time taken by Autoware.Flex for rule matching for each user instruction. Across all six experiments (three user instructions for each of the two scenarios), the maximum observed rule-matching delay is 0.77 ms for one round. This tiny delay attributes to the compact size of the rule base. At this level, the rule-matching delay can be considered negligible compared to the delays in other components within a single control cycle of the ADS.
% We also measured the time taken by Autoware.Flex for rule matching for each user instruction. Across all six experiments (three user instructions for each of the two scenarios), the maximum observed rule-matching delay was 0.77 ms. This tiny delay is attributable to the compact size of the rule base. At this level, the rule-matching delay can be considered negligible compared to the delays in other components within a single control cycle of the ADS.

% \usepackage{multirow}
\begin{table*}[t]
\scalebox{0.83}{
\begin{tabular}{|c|c|c|c|c|c|c|}
\hline
Scenario                                                                                  & Case & User instruction                                                                                  & \begin{tabular}[c]{@{}c@{}}Translated \\ AutoIR\end{tabular}                                                                                                                       & \begin{tabular}[c]{@{}c@{}}Corresponding \\ rule\end{tabular}                                                                                 & \begin{tabular}[c]{@{}c@{}}Can Autoware.Flex\\  handle?\end{tabular} & \begin{tabular}[c]{@{}c@{}}Can Autoware \\ handle?\end{tabular} \\ \hline
\multirow{3}{*}{\begin{tabular}[c]{@{}c@{}}Malfunctioning \\ traffic light\end{tabular}} & 1    & \begin{tabular}[c]{@{}c@{}}The traffic light seems \\ broken, ignore it.\end{tabular}             & \multirow{3}{*}{\begin{tabular}[c]{@{}c@{}}moduleSelect: perception\\ nodeSelect: traffic\_light\\ \_classifier\_node\\ paramSelect: use\_flag\\ configAction: FALSE\end{tabular}} & \multirow{3}{*}{\begin{tabular}[c]{@{}c@{}}Motion State: Stopped\\ Speed: = 0 m/s\\ Perceptions:\\ Traffic Light Detected\end{tabular}}       & \multirow{3}{*}{Yes}                                                 & \multirow{3}{*}{No}                                             \\ \cline{2-3}
                                                                                          & 2    & \begin{tabular}[c]{@{}c@{}}Do not follow the \\ traffic light.\end{tabular}                       &                                                                                                                                                                                    &                                                                                                                                               &                                                                      &                                                                 \\ \cline{2-3}
                                                                                          & 3    & \begin{tabular}[c]{@{}c@{}}Traffic light is crazy!\\ It is always red.\end{tabular}               &                                                                                                                                                                                    &                                                                                                                                               &                                                                      &                                                                 \\ \hline
\multirow{3}{*}{\begin{tabular}[c]{@{}c@{}}Restricted \\ lane cruising\end{tabular}}     & 1    & \begin{tabular}[c]{@{}c@{}}I want you drive on the\\ leftmost lane.\end{tabular}                  & \multirow{3}{*}{\begin{tabular}[c]{@{}c@{}}moduleSelect: planning\\ nodeSelect: mission\_planner\\ paramSelect: lane\_prefer\\ configAction: LEFT\end{tabular}}                    & \multirow{3}{*}{\begin{tabular}[c]{@{}c@{}}Motion State: Driving\\ Speed: \textless 5 m/s\\ Perceptions:\\ No Obstacle Detected\end{tabular}} & \multirow{3}{*}{Yes}                                                 & \multirow{3}{*}{No}                                             \\ \cline{2-3}
                                                                                          & 2    & \begin{tabular}[c]{@{}c@{}}Try to change to the \\ leftmost lane.\end{tabular}                    &                                                                                                                                                                                    &                                                                                                                                               &                                                                      &                                                                 \\ \cline{2-3}
                                                                                          & 3    & \begin{tabular}[c]{@{}c@{}}I wanted to get as close to \\ the left road as possible.\end{tabular} &                                                                                                                                                                                    &                                                                                                                                               &                                                                      &                                                                 \\ \hline
\end{tabular}}
\caption{The simulation results for the evaluation of instruction execution}
\label{tab:sim}
\end{table*}

\begin{figure}[t]
\centerline{\includegraphics[scale=0.24]{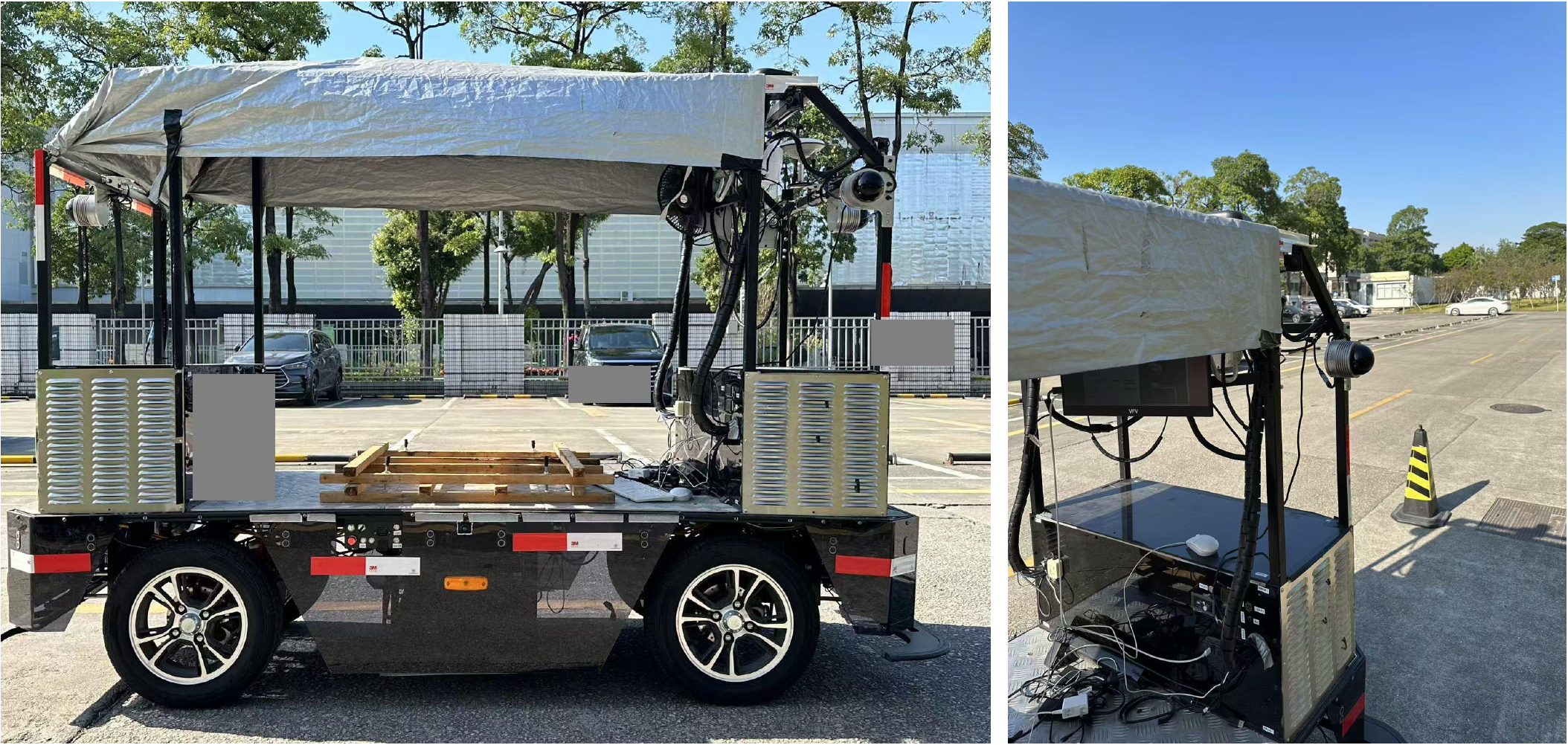}}
\caption{The prototype autonomous vehicle used in real-world evaluation}
\label{f:prototype}
\end{figure}

\subsection{Evaluation on a Real-world Autonomous Vehicle}
\label{ss:eval_auto}

To further validate Autoware.Flex in real-world environments, we develop a prototype autonomous vehicle, as shown in Fig.~\ref{f:prototype}. The prototype vehicle is built on a drive-by-wire chassis and equipped with various sensors. The on-board computer used to run the ADS features an Intel Core i9-9900K CPU clocked at 3.6 GHz, with Autoware.Flex deployed on this system.
% To further validate Autoware.Flex in real-world environments, we developed a prototype autonomous vehicle, as shown in Fig.~\ref{f:prototype}. The vehicle is built on a drive-by-wire chassis and equipped with various sensors. The on-board computer used to run the ADS features an Intel Core i9-9900K CPU clocked at 3.6 GHz, with Autoware.Flex deployed on this system.

\begin{figure*}[t]
\centerline{\includegraphics[scale=0.7]{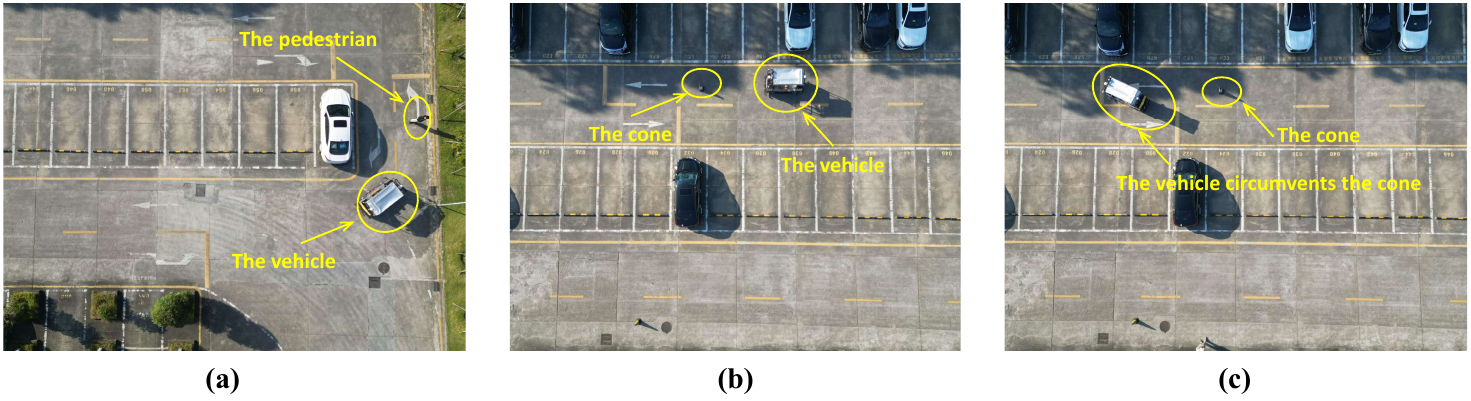}}
\caption{Scenes from the real-world experiments: (a) depicts Experiment 1, Adjusting Distance to a Pedestrian, while (b) and (c) illustrate Experiment 2, Circumventing a Traffic Cone.
}
\label{f:Real-world experiment}
\end{figure*}

We conduct multiple tests using this prototype in a real-world parking lot. Fig.~\ref{f:Real-world experiment} provides drone-captured bird’s-eye views of the vehicle in various experimental scenarios.
%We conducted multiple tests using this prototype in a real-world parking lot. Fig.~\ref{f:Real-world experiment} provides drone-captured bird’s-eye views of the vehicle in various experimental scenarios.

\textbf{\textit{Experiment 1: Adjusting Distance to a Pedestrian}}

In the first experiment, the vehicle encounter a pedestrian while driving. By default, the original ADS rule require the vehicle to stop one meter away from the pedestrian. However, if the user want to adopt a more conservative approach, they can issue an instruction to increase the stopping distance. Using user instructions, we direct the vehicle to stop farther from the pedestrian. As shown in Fig.~\ref{f:Real-world experiment} (a), the vehicle successfully stop approximately three meters away from the pedestrian, demonstrating the effectiveness of the user instruction.
% In the first experiment, the vehicle encountered a pedestrian while driving. By default, the original ADS rule required the vehicle to stop one meter away from the pedestrian. However, if the user wanted to adopt a more conservative approach, they could issue an instruction to increase the stopping distance. Using user instructions, we directed the vehicle to stop farther from the pedestrian. As shown in Fig.~\ref{f:Testbed AV}(a), the vehicle successfully stopped approximately three meters away from the pedestrian, demonstrating the effectiveness of the user instruction.

\textbf{\textit{Experiment 2: Circumventing a Traffic Cone}}

In the second experiment, the vehicle encounter a traffic cone obstructing its lane. With only one lane available in each direction, the original ADS rule causes the vehicle to stop and remain stationary. However, the user, confident that the opposite lane is clear, issue an instruction for the vehicle to use the opposite lane to bypass the cone. Fig.\ref{f:Real-world experiment} (b) shows the vehicle stop in front of the cone before the instruction is issued. After the instruction is executed, the vehicle successfully move into the opposite lane to circumvent the cone, as shown in Fig.\ref{f:Real-world experiment} (c).
% In the second experiment, the vehicle encountered a traffic cone obstructing its lane. With only one lane available in each direction, the original ADS rule caused the vehicle to stop and remain stationary. However, the user, confident that the opposite lane was clear, issued an instruction for the vehicle to use the opposite lane to bypass the cone. Fig.\ref{f:Testbed AV}(b) shows the vehicle stopped in front of the cone before the instruction was issued. After the instruction was executed, the vehicle successfully moved into the opposite lane to circumvent the cone, as shown in Fig.\ref{f:Testbed AV}(c).

\textbf{\textit{Experiment 3: Extended Stopping Time}}

In the third experiment, the vehicle again encounter a traffic cone as an obstacle. According to the original ADS rule, the vehicle will briefly stop and then seek an alternate route to bypass the obstacle. However, the user issues an instruction to extend the stopping time in front of the obstacle. This experiment involve dynamic actions that can not be effectively represented with static images, so no photos are included for this scenario.
% In the third experiment, the vehicle again encountered a traffic cone as an obstacle. According to the original ADS rule, the vehicle would briefly stop and then seek an alternate route to bypass the obstacle. However, the user issued an instruction to extend the stopping time in front of the obstacle. This experiment involved dynamic actions that could not be effectively represented with static images, so no photos are included for this scenario.

These real-world experiments strongly demonstrate Autoware.Flex’s ability to correctly interpret and execute user driving instructions, even in complex and customized scenarios. For each experiment, the corresponding AutoIR programs and matched rules are provided in Table~\ref{t:Real world results}.

\begin{table*}[t]
\scalebox{0.88}{
\begin{tabular}{|c|c|c|c|c|}
\hline
Case           & User instruction                   & Translated AutoIR                                                                                                                                                   & Corresponding rule                                                                                                        & \begin{tabular}[c]{@{}c@{}}Is user's instruction \\ executed?\end{tabular} \\ \hline
Pedestrian     & Keep a larger distance from him    & \begin{tabular}[c]{@{}c@{}}moduleSelect: planning\\ nodeSelect: behavior\_velocity\\ \_planner\_node\\ paramSelect: stop\_margin\\ configAction: 3.0\end{tabular}   & \begin{tabular}[c]{@{}c@{}}Motion State: Driving\\ Speed: \textless 5 m/s\\ Perceptions:\\ Obstacle Detected\end{tabular} & Yes                                                                        \\ \hline
Traffic Cone & Use the opposite lane to avoid it. & \begin{tabular}[c]{@{}c@{}}moduleSelect: planning\\ nodeSelect: behavior\_path\_planner\\ paramSelect: use\_opposite\_lane\\ configAction: TRUE\end{tabular}        & \begin{tabular}[c]{@{}c@{}}Motion State: Driving\\ Speed: \textless 5 m/s\\ Perceptions:\\ Obstacle Detected\end{tabular} & Yes                                                                        \\ \hline
Waiting Time & Stop for a longer time             & \begin{tabular}[c]{@{}c@{}}moduleSelect: planning\\ nodeSelect: behavior\_velocity\\ \_planner\_node\\ paramSelect: stop\_duration\\ configAction: 5.0\end{tabular} & \begin{tabular}[c]{@{}c@{}}Motion State: Stopped\\ Speed: = 0 m/s\\ Perceptions:\\ Obstacle Detected\end{tabular}         & Yes                                                                        \\ \hline
\end{tabular}}
\caption{The real-world experiment results}
\label{t:Real world results}
\end{table*}

\section{Related Work}

\vspace{0.05in}
\textbf{\textit{ - LLMs in autonomous driving}}
\vspace{0.05in}

The integration of Large Language Models (LLMs) into Autonomous Driving Systems (ADS) has attracted significant attention in recent research. Some studies explore the use of LLMs for trajectory planning in ADS. For instance, ~\cite{chen2024driving} proposes an object-level multimodal LLM architecture to enhance situational understanding in driving scenarios, while ~\cite{mao2023gpt} demonstrates a method to adapt OpenAI GPT-3.5 models into reliable motion planners for autonomous vehicles. Similarly, ~\cite{sharan2023llm} leverages the common-sense reasoning capabilities of LLMs like GPT-4 and Llama2 to improve vehicle planning. In ~\cite{fu2024drive}, the potential of LLMs to interpret driving environments in a human-like manner is explored, highlighting their reasoning, interpretation, and memory capabilities in complex scenarios. Furthermore, ~\cite{sun2024redriver} introduces a framework that utilizes LLMs to enhance decision-making processes in autonomous vehicles.
Other research focuses on employing LLMs as intermediaries for human-computer interaction. For example, ~\cite{wang2023chatgpt} presents a framework that integrates LLMs as a ``co-pilot'' for vehicles, aiming to facilitate interaction between humans and ADS. However, this approach involves direct interaction with the control model, which may introduce safety risks. Additionally, several works have explored implementing end-to-end ADS systems driven by LLMs, such as those described in ~\cite{xu2024drivegpt4}, ~\cite{shao2024lmdrive}, and ~\cite{dong2024generalizing}.

Despite their potential, LLM-driven ADS face significant challenges. As noted in \cite{cui2023drivellm} and \cite{pan2024vlp}, these systems often function as black boxes, making their decision-making processes opaque to humans. This lack of interpretability introduces considerable ethical and legal concerns. In contrast, rule-based modular ADS continue to excel in terms of safety and reliability. As discussed in \cite{yang2023llm4drive}, traditional ADS rely on well-defined rules and algorithms, ensuring predictable behavior across diverse conditions. These systems undergo rigorous testing and validation, providing consistent performance in various environments\cite{carballo2019training}\cite{zang2022winning}. Moreover, they incorporate multiple layers of redundancy and fail-safes to handle unexpected situations effectively\cite{Apollo2019}~\cite{Autoware2022}.

In this work, we leverage LLMs in a fundamentally different way. Rather than relying on LLMs for end-to-end decision-making, our primary contribution lies in integrating users' driving instructions into traditional rule-based modular ADS. This integration enables collaboration between humans and ADS during driving. Specifically, we use LLMs to translate users' driving instructions into ADS-domain-specific language. By adopting this approach, we effectively bridge the language gap between humans and ADS while retaining the modular structure of traditional ADS, thereby ensuring safety and reliability.

\vspace{0.05in}
\textbf{\textit{ - Domain-specific Languages for ADS}}
\vspace{0.05in}

Considerable research has focused on developing domain-specific languages (DSLs) for the design, testing, and functional analysis of autonomous driving systems (ADS). For instance, ~\cite{althoff2017commonroad} proposes CommonRoad, a composable road motion planning benchmark tailored for ADS design and testing. Similarly, ~\cite{fremont2019scenic} introduces Scenic, a probabilistic programming language aimed at the design and analysis of perception systems, particularly those based on machine learning.

Other notable works include ~\cite{queiroz2019geoscenario}, which develops a DSL for capturing test scenarios that reflect the complexities of real-world road traffic conditions, and ~\cite{sun2022lawbreaker}, which presents Lawbreaker, an automated framework for testing ADS compliance with real-world traffic regulations. Additionally, ~\cite{collin2020safety} designs a DSL specifically for describing traffic rules, while ~\cite{lu2024diavio} proposes a DSL for aligning real-world accident reports in natural language with violation scenarios for ADS simulation testing.

Further, ~\cite{wang2024mu} introduces $\mu$drive, a DSL designed to give users direct control over ADS. By incorporating driver preferences, $\mu$drive aims to enable safer, more stable, and more comfortable driving experiences.

While existing research emphasizes DSL design, this paper focuses on a different aspect of the problem. Although we propose AutoIR as a DSL for facilitating ADS operations, we acknowledge that other DSLs in the literature, such as $\mu$drive, could potentially serve similar purposes. The primary challenge is not the design of the DSL itself but the faithful translation of user instructions from natural language into the DSL format. This translation process is critical to bridging the gap between human intent and ADS functionality.

% % \input{sections/7-discussion}
% \input{sections/8-conclusion}
% \section{Conclusion}
% This paper identifies the task execution rate misalignment problem and proposes the Adaptive Task Execution Rate Regulation (ATER) framework to address this issue in ROS 2-based autonomous systems.
% The ATER framework operates at runtime and consists of two main components: a runtime observer, which collects and analyzes task's runtime execution events, and a task regulator, which generates optimized execution parameters to regulate task execution rates. 
% Conducted experiments under different settings on the case study system demonstrate that ATER effectively reduces message drops, conserves computational resources, and lowers end-to-end latency. 
% This approach fills the gaps in current research and provides a new perspective for enhancing the real-time performance of autonomous systems based on ROS 2.
\section{Conclusion}

Existing autonomous driving systems (ADS) independently make driving decisions based on their perception of the environment. However, these systems face significant limitations: they cannot well handle complex scenarios where environmental understanding is inadequate and are unable to incorporate human driving preferences into their decision-making processes. This paper introduces Autoware.Flex, a system that enables users to provide instructions to the ADS, guiding it toward more appropriate driving decisions. Our system addresses two key challenges: translating natural-language human instructions into an ADS-compatible format and ensuring the safe execution of these instructions within the ADS framework. Experimental results from both simulators and a real-world autonomous vehicle demonstrate the effectiveness of the proposed approach.
% Existing autonomous driving systems (ADS) independently make driving decisions based on their perception of the environment. However, these systems face significant limitations: they struggle to handle complex scenarios where environmental understanding is inadequate and are unable to incorporate human driving preferences into their decision-making processes. This paper introduced Autoware.Flex, a system that enables users to provide instructions to the ADS, guiding it toward more appropriate driving decisions. Our system addresses two key challenges: translating natural-language human instructions into an ADS-compatible format and ensuring the safe execution of these instructions within the ADS framework. Experimental results from both simulators and real-world autonomous vehicles demonstrate the effectiveness of the proposed approach.

The main contribution of this work is the novel approach to integrate human instructions into rule-based modular ADS. In the future, we aim to enhance the proposed system in the following ways: (1) developing more sophisticated AutoIR representations and leveraging advanced LLM techniques to handle complex natural language instructions; (2) enabling the system to infer the lifetime of a user instruction directly from natural language input; and (3) designing methods to automatically and incrementally expand and refine the knowledge base and rule base to support a broader range of scenarios.


\begin{acks}
Figure \ref{fig:traffic} is generated by a generative AI tool - Doubao AI. Web link: http://www.doubao.com
\end{acks}


% \bibliographystyle{ACM-Reference-Format}
% \bibliography{main}

% \appendix









\end{document}